\documentclass[conference,letterpaper]{IEEEtran}

\usepackage[utf8]{inputenc} 
\usepackage[T1]{fontenc}
\usepackage{url}
\usepackage{ifthen}
\usepackage{cite}
\usepackage[cmex10]{amsmath}
\usepackage{booktabs}
\usepackage{amssymb}
\usepackage{graphicx}
\usepackage{cuted}

\begin{document}
\title{Leveraging Conditional Mutual Information to Improve Large Language Model Fine-Tuning For Classification} 

\author{
  \IEEEauthorblockN{Thanushon Sivakaran and En-Hui Yang}
  \IEEEauthorblockA{Department of Electrical and Computer Engineering \\
                    University of Waterloo\\
                    Waterloo, Ontario\\
                    Email: \{tsivakar, ehyang\}@uwaterloo.ca}
}

\maketitle

\begin{abstract}
   Although large language models (LLMs) have demonstrated remarkable capabilities in recent years, the potential of information theory (IT) to enhance LLM development remains underexplored. This paper introduces the information theoretic principle of Conditional Mutual Information (CMI) to LLM fine-tuning for classification tasks, exploring its promise in two main ways: minimizing CMI to improve a model's standalone performance and maximizing CMI to enhance knowledge distillation (KD) for more capable student models. To apply CMI in LLM fine-tuning, we adapt the recently proposed CMI-constrained deep learning framework, which was initially developed for image classification, with some modification. By minimizing CMI during LLM fine-tuning, we achieve superior performance gains on 6 of 8 GLUE classification tasks compared to BERT. Additionally, maximizing CMI during the KD process results in significant performance improvements in 6 of 8 GLUE classification tasks compared to DistilBERT. These findings demonstrate CMI's adaptability for optimizing both standalone LLMs and student models, showcasing its potential as a robust framework for advancing LLM fine-tuning. Our work bridges the gap between information theory and LLM development, offering new insights for building high-performing language models.
\end{abstract}

\section{Introduction}
\label{sec:introduction}

Transformer-based large language models (LLMs), such as OpenAI's GPT series \cite{GPT1, GPT2, GPT3} and Google's BERT \cite{BERT}, have revolutionized the field of natural language processing (NLP) and fundamentally reshaped artificial intelligence (AI), often achieving or surpassing human-level performance in a wide range of tasks, including text generation \cite{GPT2, GPT3}, summarization \cite{TextSummarizationPretrainedEncoders, BART}, translation \cite{Attention, mBART}, and question answering \cite{BERT, XLNet}. These models are built on the transformer architecture introduced in \cite{Attention}. Their success is largely attributable to their ability to efficiently process and capture long-range dependencies and intricate contextual relationships in text. 

Behind the remarkable success of LLMs, their conceptual roots in information theory (IT) are often overlooked. Leaving their neural architectures aside, these models are nothing but computation mechanisms to establish concrete high-order Markov source models (see Section \ref{sec:BR} for details). 
In his 1948 paper \cite{Shannon} establishing the IT foundations, Shannon explored how an information source can be modeled and demonstrated several sequences of words generated randomly by first-order and second-order word-based Markov models. As the Markov order increases, the resemblance of these word sequences to ordinary English text increases quite noticeably as well. Based on this observation, Shannon then used Markov processes to model an information source.  Assuming the Markov model governing the information source is given and known, he went on to introduce key concepts like entropy and mutual information to quantify statistical uncertainty and dependencies and formalize the principles of IT. 

\begin{figure}[htbp]
  \centering
  \includegraphics[width=7.5cm]{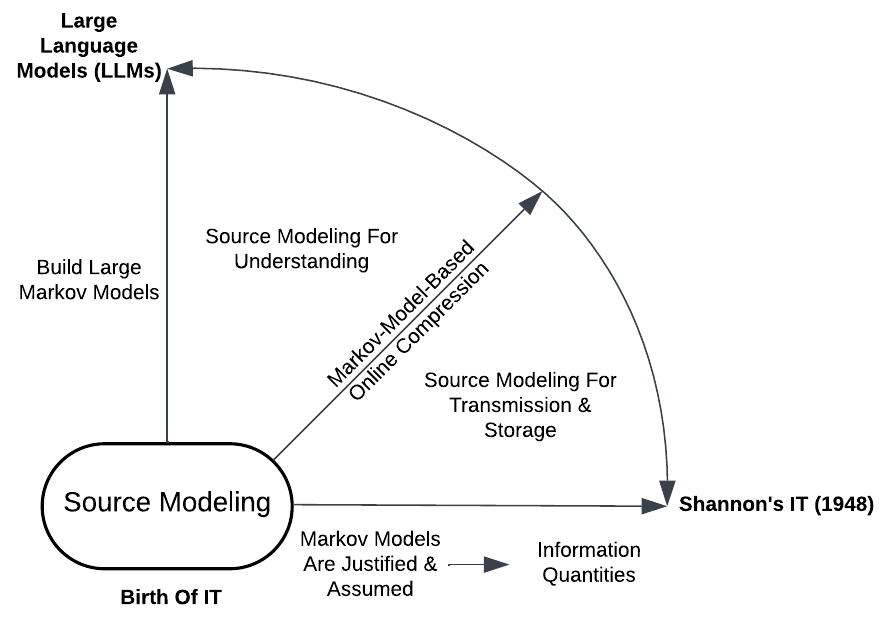}
  \caption{Relationship between LLMs and IT.}
  \label{fig:source_modeling}
\end{figure}

Shannon, nevertheless, did not establish any concrete Markov model for text. Later, during the course of tackling universal online probabilistic data compression, other researchers developed adaptive Markov source models (see \cite{yk1}, \cite{yk2}, and references therein). However, these adaptive Markov models are very small in terms of both Markov order and vocabulary size; they also lack semantic meaning. In contrast, the Markov models established by LLMs are very large in terms of both Markov order and vocabulary size, and have built-in semantic meaning. LLMs provide good solutions to the source modeling problem in IT, which has challenged information theorists even since Shannon. The relationship between LLMs and IT can be visualized according to Fig. \ref{fig:source_modeling}. To a large extent, LLMs can be considered as an extension of IT. 

How could LLMs and IT interact with each other? With reference to Fig.~\ref{fig:source_modeling}, one way is to apply LLMs to IT for better data compression. In fact, this has recently been explored in \cite{LanguageModelingIsCompression}, \cite{TransformersUniversalPredictors} to some extent. Inspired by \cite{CMI}, in this paper we instead explore the opposite direction. Specifically, we introduce the concept of Conditional Mutual Information (CMI) to LLM fine-tuning for classification tasks, exploring its promise in two main ways: minimizing CMI to improve a model's standalone performance and maximizing CMI to enhance knowledge distillation (KD) for more capable student models. To apply CMI in LLM fine-tuning, we adapt the recently proposed CMI-constrained deep learning (DL) framework \cite{CMI}, which was initially developed for image classification, with some modification. By minimizing CMI during LLM fine-tuning, we achieve superior performance gains on 6 of 8 GLUE classification tasks compared to BERT. Additionally, maximizing CMI during the KD process results in significant performance improvements in 6 of 8 GLUE classification tasks compared to DistilBERT. These findings demonstrate the potential of applying IT concepts and techniques to further advance LLM pre-training and fine-tuning. 

The rest of the paper is organized as follows. Section~\ref{sec:BR} describes LLMs in IT language and reviews LLM fine-tuning and KD. Section~\ref{sec:methodology} presents our methodology, detailing the introduction of CMI into LLM fine-tuning, its minimization during the fine-tuning process, and its maximization during the KD process. Section~\ref{sec:experiments} describes the experimental setup and discusses results. Finally, Section~\ref{sec:conclusion} concludes the paper and highlights potential directions for future research.

\section{Background and Review} 
\label{sec:BR}

Using byte pair encoding \cite{Philip}, a simple form of grammar-based coding \cite{yk,yk1, yk2}, an LLM parses a sequence of data symbols into a sequence of tokens. The set of all tokens is called a vocabulary $\cal V$. The size $|\cal V|$ of $\cal V$ is $50,257$ for GPT-3 \cite{GPT3} and $30,522 $ for BERT \cite{BERT}. Each token in $\cal V$ will be further encoded into a $d$ dimensional vector of real numbers before being fed into the LLM.

\subsection{High-Order Markov Models}  

Let \(t = [t_1, t_2, \dots, t_n]\) be a sequence of (sample) tokens. The sequence $t$ is said to be governed by a \(k\)-th order homogeneous Markov model if  the joint distribution of the  sequence \( t \)  is given by:
\begin{multline}
P(t_1, t_2, \dots, t_n) = P(t_1, t_2, \dots, t_k) \\
\times \prod_{i=k+1}^n P(t_i \mid t_{i-k}, t_{i-k+1}, \dots, t_{i-1})
\end{multline}
for any $n >k $. Here, \(P(t_1, t_2, \dots, t_k)\) is the initial distribution, which specifies the probability of the first \(k\) tokens. This can be further factorized as:
\begin{multline}
P(t_1, t_2, \dots, t_k) = P(t_1) P(t_2 \mid t_1) P(t_3 \mid t_1, t_2) \dots \\
\times P(t_k \mid t_1, t_2, \dots, t_{k-1}).
\end{multline}

Subsequent tokens are modeled by conditional probabilities based on the previous \(k\) tokens. Thus a \(k\)-th order homogeneous Markov model is uniquely specified by the first token probability distribution and the subsequent $k$ conditional distributions $ P(\cdot \mid t_1), P(\cdot \mid t_1, t_2),  \dots , P(\cdot \mid t_1, t_2, \dots, t_{k}) $.

\subsection{LLMs as High-Order Markov Source Models}  

\begin{figure}[htbp]
  \centering
  \includegraphics[width=3cm]{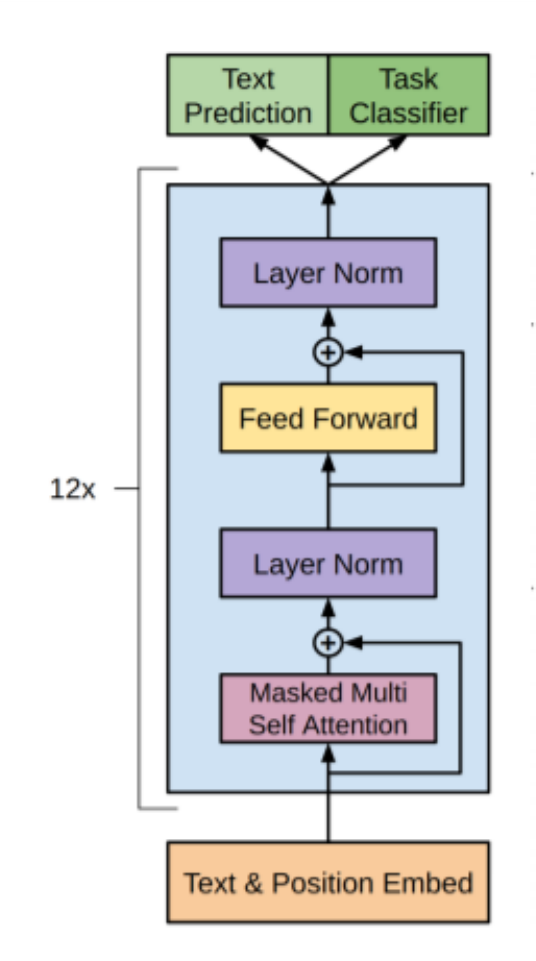}
  \caption{An illustration of a transformer-based LLM architecture (GPT-1 architecture \cite{GPT1}).}
  \label{fig:gpt1_diagram}
\end{figure}

Figure~\ref{fig:gpt1_diagram} illustrates a typical architecture of a transformer decoder-based LLM. Ignoring the final task-specific layer---whether it is for text prediction or task classification---the core structure of an LLM can be divided into three key components: (1) an input embedding layer, (2) a series of stacked transformer blocks with non-linear operations, and (3) an output layer that computes conditional probability distributions. These components are described below in terms of IT language. 

{\em Word embedding}: Encode each token $v \in {\cal V}$ into a vector $W_e (v)$ in $\mathbb{R}^d$. This can be represented by a matrix $W_e = [W_e (v_1), W_e (v_2), \dots, W_e (v_{|{\cal V}|} ) ]^T \in \mathbb{R}^{|{\cal V}| \times d} $.

{\em Position embedding}: Let $k$ be the maximum number of tokens the LLM can process in a single forward pass. $k$ is called the context length. Encode each position $1\leq i \leq k$ into a vector $W_p (i)$ in $\mathbb{R}^d$. This can be represented by a matrix $W_p = [W_p (1), W_p (2), \dots, W_p (k ) ]^T \in \mathbb{R}^{k \times d} $.

{\em Input embedding layer}: Let \(t = [t_1, t_2, \dots, t_k]\) be an input token sequence to the LLM. Each token $t_i$ is encoded into $X_0 (t_i) = W_e (t_i) + W_p (i)$. This converts $t$ into a matrix $X_0 = [ X_0 (t_1), \dots, X_0 (t_k)]^T$. 

{\em Recursive transformer blocks}: For $1 \leq l \leq L$, let $X_{l-1}$ be the output of the early layer. The $l$-th transformer block ${\cal T}_l$ converts $X_{l-1}$ into 
$X_l = [ X_l (t_1), \dots, X_l (t_k)]^T$, where $X_l (t_i) =  {\cal T}_l (X_{l-1} (t_1), \dots, X_{l-1} (t_i)) \in \mathbb{R}^d$. Note that ${\cal T}_l$ is causal. 

{\em Output layer}: For $1 \leq i \leq k$, compute the dot product of $X_L (t_i)$ with each row of $W_e$ and output the conditional distribution 
\begin{equation} \label{eq2-3}
    P(\cdot | t_1, \dots, t_i) = softmax ( X_L (t_i) W_e^T) .
\end{equation}

In the above, we omit the operation details of each transformer block ${\cal T}_l$, which involve matrix and other computation with its own parameters $\Theta_l$. An important factor is that all transformer blocks ${\cal T}_l$ are causal, unlike in encoder-based models such as BERT \cite{BERT}, which are primarily designed for downstream tasks. This, in turn, ensures that the probability distribution computed in \eqref{eq2-3} indeed can be interpreted as a conditional distribution given the present token $t_i$ and past tokens $t_1 \cdots t_{i-1}$, but not the future tokens. 

Since at the beginning, the first token is often a special token, denoted as \([Start]\) and acting as an anchor point for initializing a model, the LLM architecture described above is equivalent, mathematically, to a family of  $k$-th order homogeneous Markov source models $\{  P(\cdot | t_1, \dots, t_i; (W_e, W_p, \{\Theta_l \}_{l=1}^L)): 1 \leq i \leq k \}$ parameterized by $ \Theta = (W_e, W_p, \{\Theta_l \}_{l=1}^L)$. A specific LLM is then determined by a training process (called cross-entropy) which tries to minimize over $\Theta$,
\begin{equation} \label{eq2-4}
{\cal L}_1 ({\cal U}) =  \sum_{(u_1, \dots, u_{k+1}) \in {\cal U}} -\sum_{i=2}^{k+1} \log P(u_i | u_1, \dots, u_{i-1}; \Theta) 
\end{equation}
where $\cal U$ is a training corpus of tokens. After training, the embedding matrix $W_e$ codifies semantics into each token in the sense of \eqref{eq2-3}. The embedding dimension $d$ can be as large as  $12,288$ for GPT-3 \cite{GPT3} and $1,024$ for BERT \cite{BERT}; k is $2,048$ for GPT-3 \cite{GPT3} and $512$ for BERT \cite{BERT}.

\subsection{Fine-Tuning LLMs for Downstream Tasks}
\label{subsec:finetuning_llms}

After the LLM is pre-trained by minimizing the objective function in \eqref{eq2-4}, it can be further fine-tuned for specific downstream tasks such as classification. Let $\cal C$ be a labeled classification dataset with $C$ labels, where each sample instance consists of a sequence of tokens $ x= [x_1, \dots, x_m]$ which may include a special token, say, \([CLS]\) token, along with a label $y$. If a special token does not exist, typically the last token in the sequence is used. Pass $x$ through the pre-trained model to obtain $X_L$.  Let \(\mathbf{h}_{s}\) denote the row vector of $X_L$ corresponding to the special or last token. Feed \(\mathbf{h}_{s}\) into an added fully connected classification layer with parameter weight matrix \(\mathbf{W} \in \mathbb{R}^{d \times C}\) to predict $y$ with the following output probability distribution over \(C\) classes
\begin{equation} \label{eq2-5}
P(\cdot | x) = \text{softmax}(\mathbf{h}_{s } \mathbf{W}).
\end{equation}

The combined model with the added classification layer is further fine-tuned by minimizing over $(\Theta, \mathbf{W})$
\begin{equation}
\label{eq:final}
{\cal L}_{Final} = {\cal L}_2 ({\cal C}) + \gamma {\cal L}_1 ({\cal C}),
\end{equation}
where $\gamma$ is a hyperparameter that balances the contribution of the pre-training and supervised fine-tuning objectives, and 
\begin{equation}
\label{eq:finetuning}
{\cal L}_2 ({\cal C}) = -\sum_{(x, y) \in {\cal C}} \log P(y | x; \Theta, \mathbf{W}).
\end{equation}

\subsection{Knowledge Distillation (KD) in LLM Training}  

To address the scalability challenges posed by LLMs, KD \cite{Knowledge-Distillation-Bucilua, Knowledge-Distillation-Hinton} has emerged as a critical technique for model compression. KD transfers knowledge from a large teacher model to a smaller student model by aligning the student’s output with the teacher’s predictions. The student minimizes a combined loss
\begin{equation} \label{eq2-8}
\mathcal{L}_{\text{KD}} = (1 - \alpha) \mathcal{L}_{\text{CE}} + \alpha \mathcal{L}_{\text{KL}},
\end{equation}  
where \(\mathcal{L}_{\text{KL}} = \text{KL}(P_{\text{teacher}} \parallel P_{\text{student}})\) is the Kullback–Leibler divergence between the teacher’s and student’s softened predictions, and \(\alpha\) controls the balance between cross-entropy $ \mathcal{L}_{\text{CE}}$ and KL divergence. By distilling rich knowledge from the teacher, KD enables the student to achieve competitive performance with fewer parameters.

\section{Methodology} 
\label{sec:methodology}

We now leverage CMI to improve LLM fine-tuning for classification tasks. Specifically, we explore two complementary objectives separately: minimizing CMI to enhance the performance of standalone models and maximizing CMI to improve KD for student models.  Both approaches are inspired by CMI-constrained deep learning frameworks developed for image classification \cite{CMI, CMI-ISIT, MCMI}, but are adapted to address the unique challenges of LLM-based classification tasks.

{\em Notation}: For any integer $N >1$, let $[N] = \{ 1, 2, \dots, N \}$. For any discrete set $S$, let ${\cal P} (S)$ denote the set of all probability distributions over $S$. 

\subsection{Clusters and Centroids}

As mentioned before, for each sample instance $(x, y) \in {\cal C}$, the combined model converts the input instance $x$ first into a $d$ dimensional vector 
 \(\mathbf{h}_{s}\) (denoted as  \(\mathbf{h}_{s}^x\) and referred to as a feature vector subsequently), and then into the probability distribution $P(\cdot |x)$ in ${\cal P}([C])$ shown in \eqref{eq2-5}. For each label $y \in [C]$, let  \({\cal C}_y = \{ (x, y): (x, y ) \in {\cal C} \}\), the subset of all sample instances in $\cal C$ with label $y$, and let \(n_y = |{\cal C}_y|\) denote the cardinality of  \({\cal C}_y \). The combined model maps  \({\cal C}_y \) into a cluster of distributions $P(\cdot | x)$, $(x, y) \in {\cal C}_y$, in ${\cal P}([C])$. Refer to this cluster as the output probability distribution cluster (OPDC) corresponding to label $y$. For image classification, information quantities including CMI were introduced for these OPDCs and used to enhance the performance of DL in \cite{CMI, CMI-ISIT, MCMI}.

 For LLM-based classification tasks, however, $C$ is very small often with its value being 2 or 3, and it is not effective to use OPDCs to enhance the fune-tuning performance. To overcome this, we introduce another cluster. For each feature vector \(\mathbf{h}_{s}^x\), compute $f_x = softmax (\mathbf{h}_{s}^x) $, which is a probability distribution in ${\cal P} ([d])$. Then corresponding to each $y$, there is a cluster $\{ f_x: (x, y) \in {\cal C}_y \}$ in the distribution space ${\cal P}([d])$, referred to as the feature probability distribution cluster (FPDC) corresponding to $y$. The centroid of this FPDC is defined as
\begin{equation}
\label{eq:centroid_emp}
g_y = \frac{1}{n_y} \sum_{(x, y) \in {\cal C}_y}  f_x.
\end{equation}

Note that $d \gg C$. Next we will introduce CMI for FPDCs. 

\subsection{CMI for FPDCs}

We now want to introduce a new information quantity to measure the concentration of FPDCs. Feed an input instance $x$ into the combined model to get \(\mathbf{h}_{s}^x\), compute $f_x$ accordingly, and then use $f_x$ to randomly generate a dimension index $1 \leq z \leq d$. That is, given $x$, $z=j$ with probability $f_x (j)$. Now pick a sample instance randomly from $\cal C$; denote it by $(X, Y)$, which is random. Feed $X$ into the combined model, and let $Z$ denote the corresponding random dimension index $Z$. Then it is not hard to show that $Y \to X \to Z$ forms a Markov chain:
\begin{multline} \label{eq3-2}
P_{Z|XY} (Z=j | X=x, Y=y ) = \\
P_{Z|X} (Z=j |X =x) = f_x (j).  
\end{multline}
By the same arguments as in \cite{CMI, CMI-ISIT}, we then have for each label $y$
 \begin{equation}
\label{eq3-3}
I(X; Z |Y=y) = \frac{1}{n_y} \sum_{(x, y) \in {\cal C}_y}  KL (f_x || g_y).
\end{equation}
Thus the CMI between $X$ and $Z$ given $Y=y$ can be used to measure the concentration of FPDC corresponding to $y$. 

Averaging over all labels (or equivalently across all FPDCs), the CMI between $X$ and $Z$ given $Y$
\begin{align}
\label{eq3-4}
I(X; Z |Y) & = \frac{1}{C} \sum_{y \in [C]} \sum_{(x, y) \in {\cal C}_y}  KL (f_x || g_y) \nonumber \\
           & = \frac{1}{C}  \sum_{(x, y) \in {\cal C}}  KL (f_x || g_y)
\end{align}
then measures the average concentration of FPDCs.

\subsection{Minimizing CMI}
\label{methodMinimize}

To enhance the performance of the standalone combined model for classification, we now further include CMI $I(X; Z |Y)$ into the optimization objective. The new optimization objective is
\begin{equation} \label{eq3-5}
\mathcal{L}_{Min \;CMI} = \mathcal{L}_{Final} + \lambda I (X; Z|Y) 
\end{equation}
where \(\mathcal{L}_{Final}\) is defined in \eqref{eq:final} and \(\lambda > 0\) balances \(\mathcal{L}_{Final}\) and CMI. The combined model is now fune-tuned by a training process which tries to solve the following minimization problem
 \begin{align}
    & \min_{(\Theta, \mathbf{W})} [ \mathcal{L}_{Final} + \lambda I (X; Z|Y) ] \nonumber \\
     & = \min_{(\Theta, \mathbf{W})} [ \mathcal{L}_{Final} +  \frac{\lambda}{C}  \sum_{(x, y) \in {\cal C}}  KL (f_x || g_y) ] \nonumber \\
     & = \min_{(\Theta, \mathbf{W})} \min_{\{ G_y \}_{y \in [C]}}[ \mathcal{L}_{Final} +  \frac{\lambda}{C}  \sum_{(x, y) \in {\cal C}}  KL (f_x || G_y) ]
        \label{eq3-6}
 \end{align}
where for each $y \in [C]$, $G_y$ is a dummy distribution in ${\cal P}([d])$, and the inner minimization is achieved when $G_y = g_y$ for each $y$.

Given $ (\Theta, \mathbf{W})$ and $ \{ G_y \}_{y \in [C]}$, the objective function in \eqref{eq3-6} is now sample-instance additive and hence amenable
to parallel computation via GPU. An alternating algorithm similar to the one in \cite{CMI} can then be applied to solve the double minimization problem in \eqref{eq3-6}. 

\subsection{Maximizing CMI}
\label{methodMaximize}

To enhance the performance of distilled student models, we now maximize CMI during the fine-tuning process of the teacher model. 
Maximizing CMI would enable the teacher to encode richer contextual dependencies, improving the transfer of knowledge to the student. This approach leads to more dispersed FPDCs around their respective centroids while retaining subtle patterns in the data. The new optimization objective is
\begin{equation} \label{eq3-7}
\mathcal{L}_{Max \; CMI} = \mathcal{L}_{Final} - \lambda I(X; Z |Y),
\end{equation}
where \(\mathcal{L}_{Final}\) is defined in \eqref{eq:final} and \(\lambda > 0\) balances \(\mathcal{L}_{Final}\) and CMI. The combined model as the teacher can be fine-tuned by a training process which tries to solve the following optimization problem
\begin{align} \label{eq3-8}
    & \min_{(\Theta, \mathbf{W})} [ \mathcal{L}_{Final} - \lambda I (X; Z|Y) ].
\end{align}
To this end, an algorithm similar to the one in \cite{MCMI} can then be applied. 
A teacher trained in this manner yields predictions that preserve diverse contextual information, enhancing student performance on specialized classification tasks.

\begin{table*}[ht]
    \renewcommand{\arraystretch}{1.2}
    \caption{Performance comparison of BERT variants on dev sets of GLUE tasks. Row with * is from the DistilBERT paper \cite{DistilBERT}.}
    \label{tab:min_cmi_results}
    \centering
    \begin{tabular}{lcccccccc}
        \toprule
        Model & CoLA & MNLI & MRPC & QNLI & QQP & RTE & SST-2 & WNLI\\
        \midrule
        BERT-base* & 56.3 & \textbf{86.7} & 88.6 & \textbf{91.8} & 89.6 & 69.3 & 92.7 & 53.5 \\
        \midrule
        BERT-base (Min CMI) & \textbf{62.1} & 84.7 & \textbf{91.3} & 91.7 & \textbf{91.4} & \textbf{72.2} & \textbf{93.6} & \textbf{57.7} \\
        & +5.8\% & -2.0\% & +2.7\% & -0.1\% & +1.8\% & +2.9\% & +0.9\% & +4.2\% \\
        \bottomrule
    \end{tabular}
\end{table*}

\begin{table*}[ht]
    \renewcommand{\arraystretch}{1.2}
    \caption{Performance comparison of DistilBERT variants on dev sets of GLUE tasks. Row with * is from the DistilBERT paper \cite{DistilBERT}.}
    \label{tab:kd_results}
    \centering
    \begin{tabular}{lcccccccc}
        \toprule
        Model & CoLA & MNLI & MRPC & QNLI & QQP & RTE & SST-2 & WNLI\\
        \midrule
        DistilBERT* & 51.3 & \textbf{82.2} & 87.5 & \textbf{89.2} & 88.5 & 59.9 & 91.3 & 56.3 \\
        \midrule
        DistilBERT (Max CMI) & \textbf{51.4} & 81.6 & \textbf{88.6} & 88.3 & \textbf{90.7} & \textbf{61.7} & \textbf{91.9} & \textbf{63.4} \\
         & +0.1\% & -0.6\% & +1.1\% & -0.9\% & +2.2\% & +1.8\% & +0.6\% & +7.1\% \\
        \bottomrule
    \end{tabular}
\end{table*}

\section{Experiments}
\label{sec:experiments}

In this section, we present the experiment results on LLM fine-tuning for classification. While the proposed CMI methods with new definitions of CMI could also be applied during the pretraining phase of LLM development (where the number of centroids would correspond to the model’s vocabulary size of over 30,000), this stage entails significantly higher computational complexity. Due to resource constraints, our experiments instead focus on the fine-tuning stage, which offers a more practical approach to evaluating the effectiveness of CMI in LLM-based classification.

\subsection{Minimizing CMI for Model Optimization}
\label{subsec:minimize_CMI}

To evaluate the role of CMI minimization in optimizing LLMs for standalone performance in classification tasks, we conducted experiments on the BERT-base model \cite{BERT} for all classification tasks from the GLUE benchmark \cite{GLUE}, a suite of language understanding tasks. Each model was initialized with its pretrained weights and was fine-tuned by minimizing its CMI as explained in Section \ref{methodMinimize}. This approach will encourage the model to focus on task-relevant features while reducing noise introduced by unnecessary contextual dependencies.

The optimization process involved training the model across a range of \(\lambda\) values (0.05 to 1). Each model was trained for 5 epochs, and the best-performing epoch was selected based on the highest evaluation metric specific to each dataset. To ensure robustness, each configuration was repeated across three independent runs, and the median result was reported.

Table \ref{tab:min_cmi_results} summarizes the results of our experiments. The minimized CMI-trained BERT-base model consistently outperformed the standard BERT-base model across a majority of GLUE tasks, achieving gains on 6 out of 8 datasets. The improvements are particularly notable for CoLA and WNLI, where the minimized CMI-trained model achieved gains of 5.8\% and 4.2\%, respectively. The observed improvements can be attributed to the ability of CMI minimization to enhance the representational clarity of the model. By aligning predictions more closely with class centroids, the model achieves tighter intra-class concentration. This process reduces the influence of noisy or irrelevant contextual dependencies, allowing the model to prioritize features that are most critical for the primary task objectives.

\subsection{Maximizing CMI for Student Models}
\label{subsec:maximize_CMI}

To investigate the effectiveness of maximizing CMI in facilitating knowledge transfer from teacher to student models, we conducted experiments designed to fine-tune teacher models by maximizing their CMI before applying KD as explained in Section \ref{methodMaximize}. For these experiments, we followed a similar procedure to the one described in Section \ref{subsec:minimize_CMI}. The BERT-base model \cite{BERT} was used as our teacher model which was initialized with its pretrained weights and fine-tuned by maximizing its CMI. For fine-tuning, we explored a range of \(\lambda\) values, from 0.05 to 1, to balance the log-likelihood and CMI terms. Each model was trained for 5 epochs, and the best-performing epoch for each run was selected based on the highest evaluation metric specific to the dataset. To ensure robustness, each \(\lambda\) configuration was repeated across three independent runs. 

To distill knowledge to the student model, we selected the teacher model that exhibited the highest evaluation metric-to-CMI ratio among all the \(\lambda\) configurations. Notably, most of the chosen teacher models corresponded to \(\lambda\) values between 0.05 and 0.40, where the trade-off between contextual richness and the evaluation metric was most effective.

The student model used was DistilBERT \cite{DistilBERT}, which was initialized by copying every other layer from the teacher model, as described in the original DistilBERT paper. For distillation, we employed the traditional KD framework \cite{Knowledge-Distillation-Hinton}, with \(\alpha\) values ranging from 0.05 to 0.9 and temperature values from 1 to 5. Each student was trained for 10 epochs, with the best-performing model selected based on the highest evaluation metric specific to the dataset from the median of three runs. These results can be seen in Table~\ref{tab:kd_results}.

The students distilled from maximized CMI-trained teachers demonstrate substantial improvements over those trained using the original DistilBERT process, achieving greater results on 6 out of 8 GLUE classification tasks. Notably, on the WNLI and QQP datasets, the gains are particulary significant, with improvements of 7.1\% and 2.2\% respectively. These results emphasize the advantages of maximizing CMI in enabling teachers to encode and transfer richer contextual relationships to their students. By capturing broader patterns and preserving nuanced dependencies in the data, the maximized CMI-trained teachers enhance the student's ability to perform effectively for specialized classification tasks.

\section{Conclusion}
\label{sec:conclusion}

In this paper, we have connected LLMs to IT by describing LLMs in terms of IT language and demonstrating that transformer decoder-based LLMs are mathematically equivalent to computable high-order homogeneous Markov source models. We have then leveraged CMI to enhance LLM fine-tuning by adapting the CMI-constrained deep learning framework, modified for LLM-based classification tasks. Minimizing CMI during fine-tuning improves standalone performance by sharpening intra-class concentration, effectively aligning predictions with their respective centroids. Meanwhile, maximizing CMI during knowledge distillation facilitates the transfer of richer contextual relationships from teacher to student models, enabling the latter to achieve superior task-specific performance.

Future work could extend the application of CMI beyond classification tasks to areas such as text generation, multi-modal learning, etc. Exploring these directions could uncover broader applications of information-theoretic principles in advancing NLP and beyond.

\clearpage

\bibliographystyle{IEEEtran}
\bibliography{paper}

\end{document}